%% file: MLOps_perf_pred.tex
\documentclass[conference]{IEEEtran}
\IEEEoverridecommandlockouts
\usepackage{cite}
\usepackage{amsmath,amssymb,amsfonts}
\usepackage{algorithmic}
\usepackage{graphicx}
\usepackage{textcomp}
\usepackage{xcolor}

\usepackage{url}
\usepackage{verbatim}
\usepackage{multirow}
\usepackage{multicol}
\usepackage{enumitem}
\usepackage{epsfig}
\usepackage{subfigure}
\graphicspath{{./FIG/}}
\usepackage{balance}
\usepackage{multicol}
\usepackage{booktabs}

\def\BibTeX{{\rm B\kern-.05em{\sc i\kern-.025em b}\kern-.08em
    T\kern-.1667em\lower.7ex\hbox{E}\kern-.125emX}}
\begin{document}

\title{Data Complexity-aware Deep Model Performance Forecasting
\thanks{The research is partially supported by the Ministry of Science and
Technology grant MOST 109-2221-E-011-127-MY3, and National
Science and Technology Council grants NSTC 112-2221-E-011-111, NSTC 113-2221-E-011-119, NSTC 112-2634-F-011-002-MBK, and NSTC 113-2634-F-011-002-MBK, Fujian Natural Science Foundation (2025J011063), and Wuyi University Talent Introduction Research Startup Project (YJ202512).}
}

\author{\IEEEauthorblockN{Yen-Chia Chen, \hspace{0.1in} Hsing-Kuo Pao}
\IEEEauthorblockA{\textit{Dept. of Computer Science and Information Engineering} \\
\textit{National Taiwan University of Science and Technology}\\
Taipei, Taiwan 106 \\
charlisyenchiachen@gmail.com, \hspace{0.1in} pao@mail.ntust.edu.tw}
\and
\IEEEauthorblockN{Hanjuan Huang}
\IEEEauthorblockA{\textit{College of Mechanical and Electrical Engineering} \\
\textit{WUYI University}\\
Wuyishan, 354300, China \\
huanghanjuan@wuyiu.edu.cn}
}

\maketitle

\input{000_abstract}
\input{100_intro}
\input{200_related}
\input{300_method}

\input{400_results}

\input{500_conclusion}

\balance
\bibliographystyle{IEEEtran} 
\bibliography{BIB/perf_pred,BIB/IBref,BIB/ml,BIB/cs,BIB/tm,BIB/dcm, BIB/dataset}
\balance

\end{document}

%% file: 000_abstract.tex
\begin{abstract}
Deep learning models are widely used across computer vision and other domains.
When working on the model induction, selecting the right architecture
for a given dataset often relies on repetitive trial-and-error procedures.
This procedure is time-consuming, resource-intensive, and difficult to automate.
While previous work has explored performance prediction using partial training or complex simulations,
these methods often require significant computational overhead or lack generalizability.
In this work, we propose an alternative approach: a lightweight, two-stage framework that
can estimate model performance before training given the understanding of the dataset
and the focused deep model structures.
The first stage predicts a baseline based on the analysis of some measurable properties of the dataset,
while the second stage adjusts the estimation with additional information
on the model's architectural and hyperparameter details.
The setup allows the framework to generalize across datasets and model types.
Moreover, we find that some of the underlying features used
for prediction—such as dataset variance—can offer practical guidance for model selection,
and can serve as early indicators of data quality.
As a result, the framework can be used not only to forecast model performance,
but also to guide architecture choices, inform necessary preprocessing procedures,
and detect potentially problematic datasets before training begins.
\end{abstract}


%% file: 100_intro.tex
\section{Introduction}
\label{sec:introduction}

Training deep learning models often takes a tremendous amount of computing resources.
The typical cycle of designing, tuning, and validating models can be slow and expensive,
which makes it hard to deploy to diverse scenarios.
The aspect of green computing concerns us about how to accomplish deep learning tasks with
as minimal energy usage as possible. As a result, having a wise energy-aware plan in advance
when dealing with deep learning tasks is especially important.
In this work, we aim to have the ability to estimate how well a specific model may perform
\textit{before} the training for convergence. Based on such a prophetic action,
we have the choice of proceeding or not proceeding with the complete
and precise deep induction procedure.
Moreover, the estimated performance may be already good enough
when only approximated performance is needed to justify the model's ability.
Overall, the proposed method serves as a beneficial pre-run procedure
to save unnecessary computation, in nowadays when machine learning operations (MLOps)
has become a focused discipline.

The focused problem is to ask for a hint about the performance of a deep model
before the formal training of the model.
The hint may suggest that the trained model may or may not work well,
then, leading to several action plans:
(i) the model may work well and we would like to know the precise model performance
after a formal training of the model;
(ii) we understand the model's approximated performance up to certain degree
and the approximation is good enough to make further decisions,
such as good enough to evaluate between the models that own various settings or parameters;
(iii) the model may not work well or work below our expectation
and we should avoid the use of such deep model.
The computation that is needed to perform the complete training of the deep learner is saved.

A practical solution for the aforementioned problem must satisfy four requirements simultaneously: (i) operate strictly \textit{before} training, (ii) be computationally lightweight, (iii) generalize across datasets and domains, and (iv) be interpretable enough to inform design choices.
Existing approaches may miss one or more of these requirements.
The learning curve–based methods still require partial training runs, which go against the idea of avoiding early computation~\cite{domhan2015speeding}. On the other hand, 
some other approaches rely on complex surrogates, such as Graph Neural Networks (GNNs), that can take as much time and memory as training the actual model~\cite{gao_runtime_2023}.
In addition, many methods act like black boxes, offering limited explanations of why a model might succeed or fail. We lack the possibility from further improving
or revising the training strategy.
For instance, White et al.~\cite{white2021bananas} proposed BANANAS, a neural predictor that accelerates NAS through path-based encodings, but it remains essentially a predictive surrogate with limited interpretability regarding why certain architectures may perform well.

In this paper, we propose an efficient two-stage framework to predict model's performance
before its formal training given a dataset and the chosen learning model.
The novel design in the framework is to separate the predictive problem into two stages.
The first stage is to estimate how difficult the dataset is.
The more difficult the dataset is, the harder for a deep model to
output accurate prediction on the data.
The data difficulty may imply highly overlapped regions between data that own different labels
in the supervised sense, or between different groups of data in the unsupervised sense.
Other difficulty may have non-smooth decision boundaries between
data that own different characteristics, or data may contain latent features.
The data difficulty surely can influence the performance of whatever chosen deep model
that may be applied to the dataset.
In the first stage, we use a set of predefined complexity measures to
estimate a baseline performance level based on the dataset characteristics alone.
 
The second stage is to confirm the model performance prediction result once
the deep model architecture is decided. Given the baseline prediction from the first stage,
we may have certain deviation for different chosen deep models.
Therefore, we confirm how much offset from the baseline prediction is
once the deep model is chosen or the deep architectures are decided.
Compared to the one stage designation, the proposed decomposed framework
reflects two types of factors that may have impact on the model's performance.
The dataset difficulty governs the first-order performance and it well captures
the dataset characteristics that may influence the deep model performance,
while the model's architecture and hyperparameters may contribute the higher-order,
and conditional deviations that should be handled by a nonlinear learner.
We describe these components and their empirical support in Section~\ref{sec:exploratory_analysis}.

We summarize the key properties and contributions in this work as follows:
\begin{itemize}
    \item {\em A Two-Stage Predictive Framework:} We introduce a general method to
    predict the {\em deep model performance} for green computing with a two-stage design.
    Both the dataset characteristics and the domain the dataset belongs to
    shall be considered in the prediction.
    The proposed method is tested through Leave-One-Dataset-Out (LODO) and Leave-One-Domain-Out (LODM) validation.
    
    \item {\em Data-Guided Architecture Selection:} We find that one data metric  ({\em Variance Mean}) is strongly linked to model depth, offering a simple rule of thumb for choosing model size based on dataset properties.
    The data difficulty measure is known before the formal deep training
    and we can use the important cues and the early-prediction to decide what could be
    deep model structures.
    
    \item {\em Early Diagnostic Signal for Data Quality:} One of the learned components ({\em PC6}) appears to identify datasets with likely bias or variance issues.
    It may help to catch problems before training starts.
    It means that we can perform certain data cleaning or preprocesing procedures
    to have robust training after the cleaned data are prepared.
    
    \item {\em Low Data Requirement:} The method only needs a small portion of the dataset (about 16\%) to estimate its difficulty metrics, which makes it efficient and easy to
    be deployed practically even when resources are limited.
    Moreover, we may have the proposed method run before a complete dataset is compiled.
    That means, we may execute the proposed prediction before we acquire or collect
    the complete dataset. It can save a lot of efforts on the data collection procedure.
\end{itemize}

The rest of this paper is organized as follows. Right after the introduction,
we review related work for the proposed method in Section~\ref{sec:related_work}.
In Section~\ref{sec:methodology}, we introduce the proposed method and
the rationale that may support the designation of the method.
It is followed by its evaluations and discussion in Section~\ref{sec:results_discussion}.
In the end, we conclude the work in Section~\ref{sec:conclusion_future_work}.

%% file: 200_related.tex
\section{Related Work}
\label{sec:related_work}

Performance prediction for deep learning models has received increasing attention due to the high computational cost of training and the need for efficient resource planning. Prior work spans analytical modeling, data-driven prediction, and meta-feature-based approaches.

Qi et al.~\cite{qi_paleo_iclr_2017} proposed \textit{PALEO}, an analytical performance model that estimates the execution time of deep neural networks by decomposing it into computation and communication components. PALEO models per-layer operations based on architecture specifications, hardware capabilities, and communication strategies. It supports scalability analysis across different parallelization schemes and enables performance estimation without actual execution. While effective for runtime estimation, PALEO focuses on system-level efficiency rather than model accuracy prediction.

Justus et al.~\cite{justus_pred_comp_bigdata_2018} addressed training time prediction by modeling the execution time of individual network components. Their method leverages a deep network to learn execution time mappings, which are then aggregated to obtain end-to-end runtime. This approach captures non-linear factors such as memory bottlenecks and hardware-specific inefficiencies. However, it remains limited to temporal prediction and does not consider dataset properties or learning dynamics.

Gao et al.~\cite{gao_runtime_2023} discussed how to use graph neural network to predict the performance of deep learning models. One of the major concerns is due to the time spent on graph neural networks may not be shorter than that could be spent on deep learners. In the proposed method, we instead adopt random forests for the prediction, which makes more sense as we can expect a quick understanding on the deep learning performance before the deep learning training can be finished.

While many existing methods focus on estimating runtime or system cost, others aim to predict model accuracy directly. One common approach is learning curve extrapolation. For example, Domhan et al.~\cite{domhan2015speeding} showed that fitting simple functions (such as power laws) to early validation results can help estimate final accuracy, which allows early stopping of less promising models. Building on this idea, Freeze-Thaw Bayesian Optimization~\cite{swersky2014freeze} uses Gaussian processes to model partial learning curves and improve resource allocation. Although useful in practice, these methods still rely on partial training and can be affected by random variation and hyperparameter schedules.

To avoid the need for partial training, some researchers have turned to meta-learning. In this setting, accuracy prediction is treated as a supervised regression problem, using results from past experiments across different datasets and models. These methods use meta-features such as dataset-level statistics, classifier outputs~\cite{pfahringer2000meta}, or embeddings like Dataset2Vec~\cite{jomaa2021dataset2vec}. While flexible, meta-learning often requires a large amount of prior data and complex preprocessing to extract the input features.

As a simpler alternative, Data Complexity Measures (DCMs) have long been used in classical machine learning to describe how difficult a classification problem is~\cite{ho2002complexity, lorena2019complex}. These measures capture properties like class overlap, decision boundary shape, and the structure of the feature space. Although originally developed for algorithm selection and task comparison, their use in predicting deep learning performance has received limited attention.

In this work, we explore how DCMs can be combined with basic model descriptors to estimate model accuracy before training. Our method is trained on historical performance data, learning how dataset complexity relates to model behavior, but it does not require learning curves or task-specific metadata at inference time. The goal is to support early decision-making about models and datasets, especially when computational resources are limited.

\label{sec:past}

%% file: 300_method.tex
\section{Methodology}
\label{sec:methodology}

\subsection{Problem Setting and Design Principles}
\label{sec:setting}
Given a dataset $D$ and a model configuration $m$ (architecture family and hyperparameters), let $A(m,D)$ denote the converged test accuracy. We aim to design a pre-training predictor $g$ that outputs $\hat{A} = g(D,m)$ under constraints of  minimal computational overhead and applicability across heterogeneous datasets and domains.

Two principles guide our design: (i) {\em Decoupling}: decompose performance into a dataset-driven \emph{baseline} and an architecture-conditional \emph{offset}, $A(m,D) \approx A_{\text{base}}(D) + O(m,D)$; (ii) {\em Parsimony then expressivity}: use a linear model for the first-order baseline and a nonlinear regressor for higher-order, conditional deviations.

\begin{figure}[htbp]
    \centering
    \includegraphics[width=0.9\linewidth]{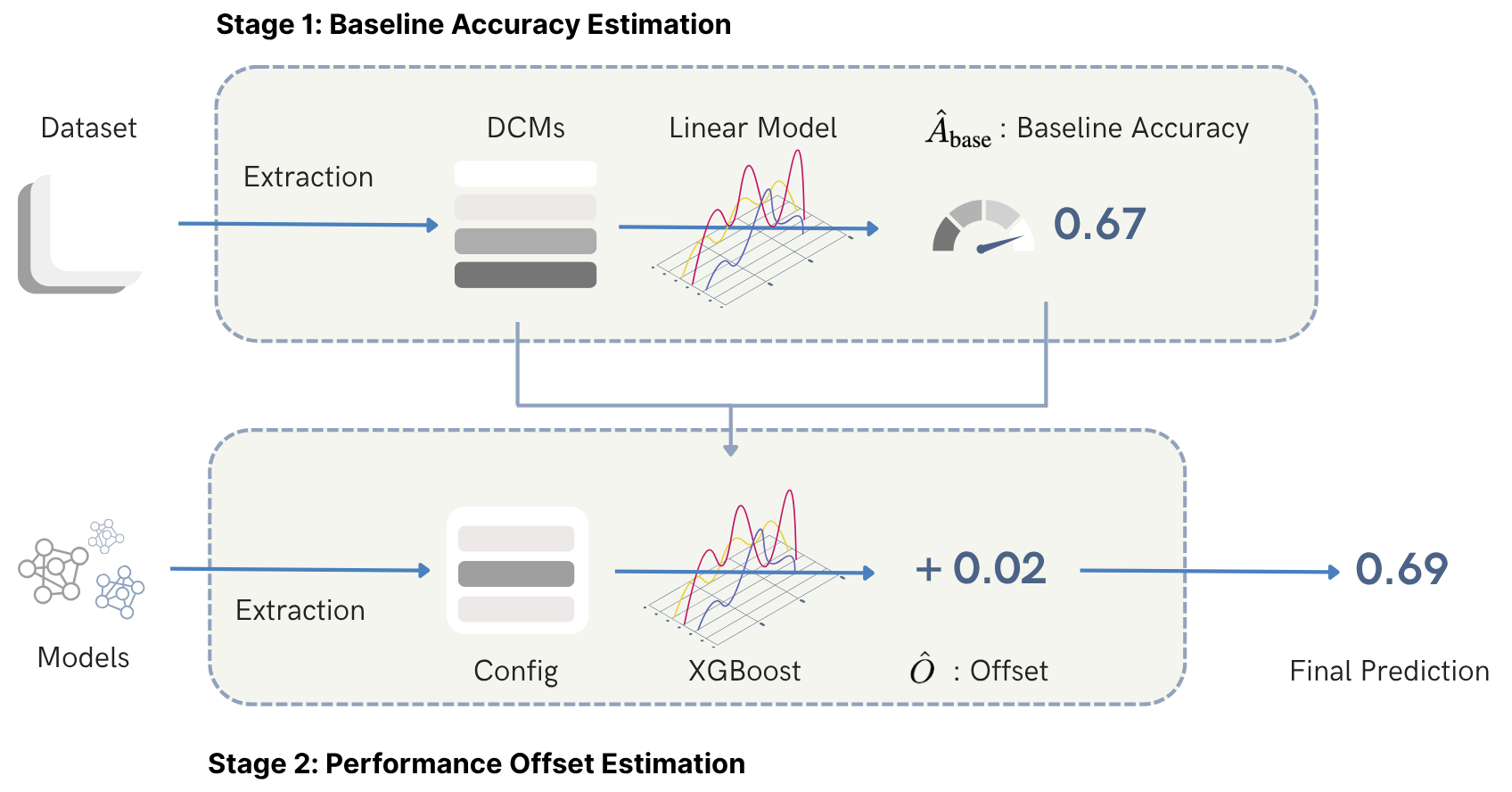}
    \caption{Two-stage prediction framework.}
    \label{fig:framework}
\end{figure}

\subsection{Overall Framework Architecture}
\label{sec:framework_architecture}

Our framework instantiates the above principles in a two-stage pipeline; we estimate a dataset-intrinsic baseline accuracy, $\hat{A}$, from a compact complexity basis obtained via principal component analysis (PCA) of data complexity measures (DCMs). In stage~2, we predict a performance offset, $\hat{O}$, conditioned on both model architecture descriptors and the baseline. PCA provides the basis for complexity features; empirical support for this decomposition appears in Section~\ref{sec:exploratory_analysis}. This modular design—Stage 1 modeling “problem difficulty” and Stage 2 modeling “solution quality”—improves predictive accuracy, maintains interpretability, and supports deployment in resource-constrained MLOps settings.

\begin{table*}[htbp]
\centering
\caption{Summary of Data Complexity Measures (DCMs).}
\label{tab:dcm_descriptions}
\begin{tabular}{@{}lll@{}}
\toprule
{\em Category} & {\em Measure Name} & {\em Description} \\
\midrule
Feature-based & Covariance Mean & Average feature-pair covariance \\
 & Variance Mean & Average feature variance \\
 & Max Fisher's Ratio & Highest linear separability per feature \\
 & Overlap Region & Feature range overlap between classes \\
 & Max Feature Efficiency & Best feature’s discriminative power \\
\midrule
Linearity & Linear Classifier Error & Error of linear classifier \\
\midrule
Neighborhood & NN Distance Ratio & Ratio of intra/inter-class distances \\
 & k-NN Error Rate & Error rate of 3-NN classifier \\
 & NN Non-linearity & Error rate with interpolated points \\
\midrule
Dimensionality & Raw Feature Count & Original feature space dimension \\
 & PCA Components & Components retaining 95\% variance \\
 & PCA Retention Ratio & Effective-to-original dimension ratio \\
\midrule
Class Balance & Class Entropy & Entropy of class distribution \\
 & Imbalance Ratio & Minority-to-majority class size ratio \\
\bottomrule
\end{tabular}
\end{table*}


\subsubsection{Stage 1: Baseline Accuracy Estimation}

Stage~1 estimates a baseline accuracy ($\hat{A}_{\text{base}}$) intrinsic to each dataset, independent of model specifics. We utilize Ordinary Least Squares (OLS) regression on principal components (PCs) derived from DCMs:
\begin{equation}
    \hat{A}_{\text{base}} = \beta_0 + \sum_{i=1}^{N}\beta_i \cdot \text{PC}_i + \varepsilon,
    \label{eq:stage1}
\end{equation}
where $\beta_i$ are regression coefficients, $\varepsilon$ is residual error, and $\text{PC}_i$ are selected PCs ($N=7$).


\subsubsection{Stage 2: Performance Offset Estimation}

The second stage models the performance deviation, $\hat{O}$, from the baseline, which arises from the specific choice of model architecture and hyperparameters. This offset captures the complex, non-linear interactions between a model’s design and a dataset’s characteristics. For this regression task, we employ XGBoost, a gradient boosting model known for its effectiveness in modeling structured data.

The offset is formally predicted by:
\begin{equation}
    \hat{O} = f_{\text{XGB}}(\mathbf{x}_{\text{DCM}}, \mathbf{x}_{\text{arch}}, \hat{A}_{\text{base}}),
    \label{eq:stage2}
\end{equation}
where $\mathbf{x}_{\text{DCM}}$ are the same PCA-transformed complexity features, and $\mathbf{x}_{\text{arch}}$ is a vector of architecture descriptors (e.g., depth, filter count, learning rate).

Notably, we include the Stage~1 prediction, $\hat{A}_{\text{base}}$, as an input to the Stage~2 model. This allows the XGBoost model to learn conditional effects; for example, it can learn that a high dropout rate might be beneficial for an ``easy'' dataset (high $\hat{A}_{\text{base}}$) but detrimental for a ``hard'' one (low $\hat{A}_{\text{base}}$).

The final predicted accuracy, $\hat{A}_{\text{final}}$, is the sum of the baseline and the predicted offset:
\begin{equation}
    \hat{A}_{\text{final}} = \hat{A}_{\text{base}} + \hat{O}.
    \label{eq:final_pred}
\end{equation}
This formulation provides clear, modular interpretability: $\hat{A}_{\text{base}}$ quantifies the ``problem difficulty'', while $\hat{O}$ quantifies the ``solution quality'' for that specific problem.


\subsection{Rationale for Model Selection}

The selection of predictive models for each stage of our framework is deliberate, tailored to the distinct nature of the relationships being modeled. Stage~1 addresses a predominantly linear problem, while Stage~2 tackles complex, non-linear interactions.

\paragraph{Stage 1: Linear Model for Baseline Prediction}
The objective of Stage~1 is to capture the relationship between a dataset's intrinsic complexity and its expected (mean) performance. Our empirical analysis across all training configurations reveals a strong, approximately linear correlation between dataset identity and achievable accuracy. As illustrated by the performance distributions in our preliminary analysis (detailed in Section~\ref{sec:exploratory_analysis}), simpler datasets like MNIST consistently yield high accuracies with low variance, regardless of model configuration. Conversely, complex datasets like CIFAR-100 exhibit significantly lower and more dispersed accuracies.

This observation suggests that intrinsic dataset difficulty is the dominant, first-order determinant of performance. Given this predominantly linear relationship, we employ Ordinary Least Squares (OLS) regression for Stage~1. OLS provides a parsimonious, interpretable, and computationally efficient model to establish the baseline accuracy ($\hat{A}_{\text{base}}$) from the principal components of data complexity.

\paragraph{Stage 2: Non-Linear Model for Offset Prediction}
The aim of Stage~2 is to predict the performance offset ($\hat{O}$), which quantifies the extent to which actual performance deviates from the baseline as a result of specific choices in model architecture and training configuration. This offset represents interactions that are challenging to model directly; for instance, a learning rate suitable for one dataset may not perform well for another, influenced by factors such as data complexity. Given the non-linear nature of these relationships, which depend on both model and dataset characteristics, we employ XGBoost, a gradient boosting method capable of capturing higher-order feature interactions. The model receives as input the dataset complexity measures, model descriptors, and the baseline estimate from Stage~1 to predict the final offset.


\subsection{Feature Engineering and Selection}

To enable performance forecasting prior to training, our framework's inputs are divided into two distinct categories, aligning with its two-stage design: data-inherent complexity features and model-specific architectural descriptors.

\subsubsection{Data Complexity Measures (DCMs)}
Data Complexity Measures (DCMs) are statistical metrics that quantify the intrinsic difficulty of a classification task based on the geometric and statistical properties of the data~\cite{ho2002complexity, lorena2019complex}. 

In this work, we curate a comprehensive set of established DCMs and further propose two global descriptors: \textit{Covariance Mean} and \textit{Variance Mean}. These descriptors summarize inter-feature correlation and overall feature dispersion, respectively, and are computed directly from raw input data without any training. The full set of measures, detailed in Table~\ref{tab:dcm_descriptions}, spans five aspects of data structure: feature distribution, linearity, neighborhood geometry, dimensionality, and class balance. These DCMs serve as the exclusive inputs for Stage~1 of our framework.


\subsubsection{Model Architecture Descriptors}
To model the influence of architectural choices, we define a set of lightweight descriptors that abstract the target model's structure. These are divided into two types:
\begin{itemize}
    \item {\em Continuous Descriptors:} Key hyperparameters that define model capacity and training dynamics, including the number of convolutional filters, number of dense units, dropout rate, and learning rate.
    \item {\em Categorical Descriptors:} The architectural family (e.g., LeNet, VGG, ResNet), encoded as a categorical variable. This captures high-level design paradigms like residual connections or network depth without needing to parse a full computation graph.
\end{itemize}
These computationally inexpensive descriptors, combined with the DCMs, form the complete input set for the Stage~2 offset prediction model.


%% file: 400_results.tex
\section{Results and Discussion}
\label{sec:results_discussion}

This section validates and analyzes the proposed framework in six parts:
(1) \textit{Experimental Setup} — datasets, architectures, and hyperparameters used to generate performance data;
(2) \textit{Exploratory Analysis} — empirical motivation for the two-stage design;
(3) \textit{Performance Validation} — predictive accuracy under in-distribution, LODO, and cross-domain settings;
(4) \textit{Ablation Study} — necessity of decoupling baseline and offset;
(5) \textit{Practical Considerations and Diagnostics} — sample efficiency and error-as-signal diagnostics; and
(6) \textit{Generalization Beyond Image Data} — applicability outside vision tasks.

\subsection{Experimental Setup}

We detail the empirical setup used to generate the training signals for our predictors, covering the datasets, model architectures, and hyperparameters. The selection emphasizes diversity in domain, scale, and complexity to enable robust assessment of the DCM-based two-stage framework.

\subsubsection{Datasets}

We employ seven publicly available image-classification datasets spanning handwriting, objects, and medical imaging. They differ in numbers of classes, sample sizes, resolutions, and visual complexity, providing a comprehensive testbed for evaluating generalizability. Table~\ref{tab:datasets} summarizes their characteristics. We follow the standard train/test splits for each dataset whenever available.

\begin{table}[htb]
\centering
\caption{Overview of Datasets Used in the Study}
\label{tab:datasets}
\begin{tabular}{lllll}
\toprule
\textbf{Dataset} & \textbf{Domain} & \textbf{Classes} & \textbf{Train} & \textbf{Test} \\
\midrule
MNIST & Handwriting & 10 & 60,000 & 10,000 \\
notMNIST & Handwriting & 10 & 18,726 & 1,872 \\
Fashion-MNIST & Objects & 10 & 60,000 & 10,000 \\
CIFAR-10 & Objects & 10 & 50,000 & 10,000 \\
CIFAR-100 & Objects & 100 & 50,000 & 10,000 \\
PathMNIST & Medical & 9 & 89,996 & 7,180 \\
BloodMNIST & Medical & 8 & 11,959 & 3,421 \\
\bottomrule
\end{tabular}
\end{table}

All images are standardized per channel using dataset-specific statistics. Predefined training/testing splits were strictly followed for both DCM computation and model performance evaluation to avoid leakage.

\subsubsection{Model Architectures}
To examine interactions between dataset complexity and model design, we use three representative deep learning models with increasing depth and architectural complexity: LeNet~\cite{lecun2002gradient}, VGG~\cite{simonyan2014very}, and ResNet~\cite{he2016deep}. These models exemplify distinct design paradigms—from shallow early architectures to deep networks with advanced features such as residual connections. Table~\ref{tab:architectures} summarizes the selected model variants.

\begin{table}[htb]
\centering
\caption{Selected Model Architectures}
\label{tab:architectures}
\begin{tabular}{lll}
\toprule
\textbf{Family} & \textbf{Variant} & \textbf{Description} \\
\midrule
LeNet & LeNet-5 & 2 Conv + 3 FC layers \\
VGG & VGG-16 & 13 Conv + 3 FC layers \\
ResNet & ResNet-18 & 17 Conv (Residual) + 1 FC layer \\
\bottomrule
\end{tabular}
\end{table}

All models were trained for a fixed number of epochs without early stopping. This ensures a wide spectrum of final accuracy values, facilitating robust training of our predictive framework.

\subsubsection{Hyperparameter Space}
To induce performance diversity across identical architectures, we systematically varied four key hyperparameters: number of convolutional filters, dense layer width, dropout rate, and learning rate. The values explored for each parameter are summarized in Table~\ref{tab:hyperparams}.

\begin{table}[htb]
\centering
\caption{Explored Hyperparameter Values}
\label{tab:hyperparams}
\begin{tabular}{ll}
\toprule
\textbf{Hyperparameter} & \textbf{Values} \\
\midrule
Filters & \{8, 16, 32, 64\} \\
Dense units & \{64, 128, 256\} \\
Dropout rate & \{0.0, 0.25, 0.5\} \\
Learning rate & \{0.001, 0.0005, 0.0001\} \\
\bottomrule
\end{tabular}
\end{table}

We adopted a full factorial design across all dataset-architecture pairs, resulting in a diverse collection of training runs. Each configuration yielded a unique test accuracy, which served as the target variable for learning the performance prediction model.

\subsection{Exploratory Analysis: Motivating the Two-Stage Design}
\label{sec:exploratory_analysis}

\subsubsection{The Dominant Role of Dataset Difficulty}

\begin{figure}[htbp]
  \centering
  \includegraphics[width=0.95\linewidth]{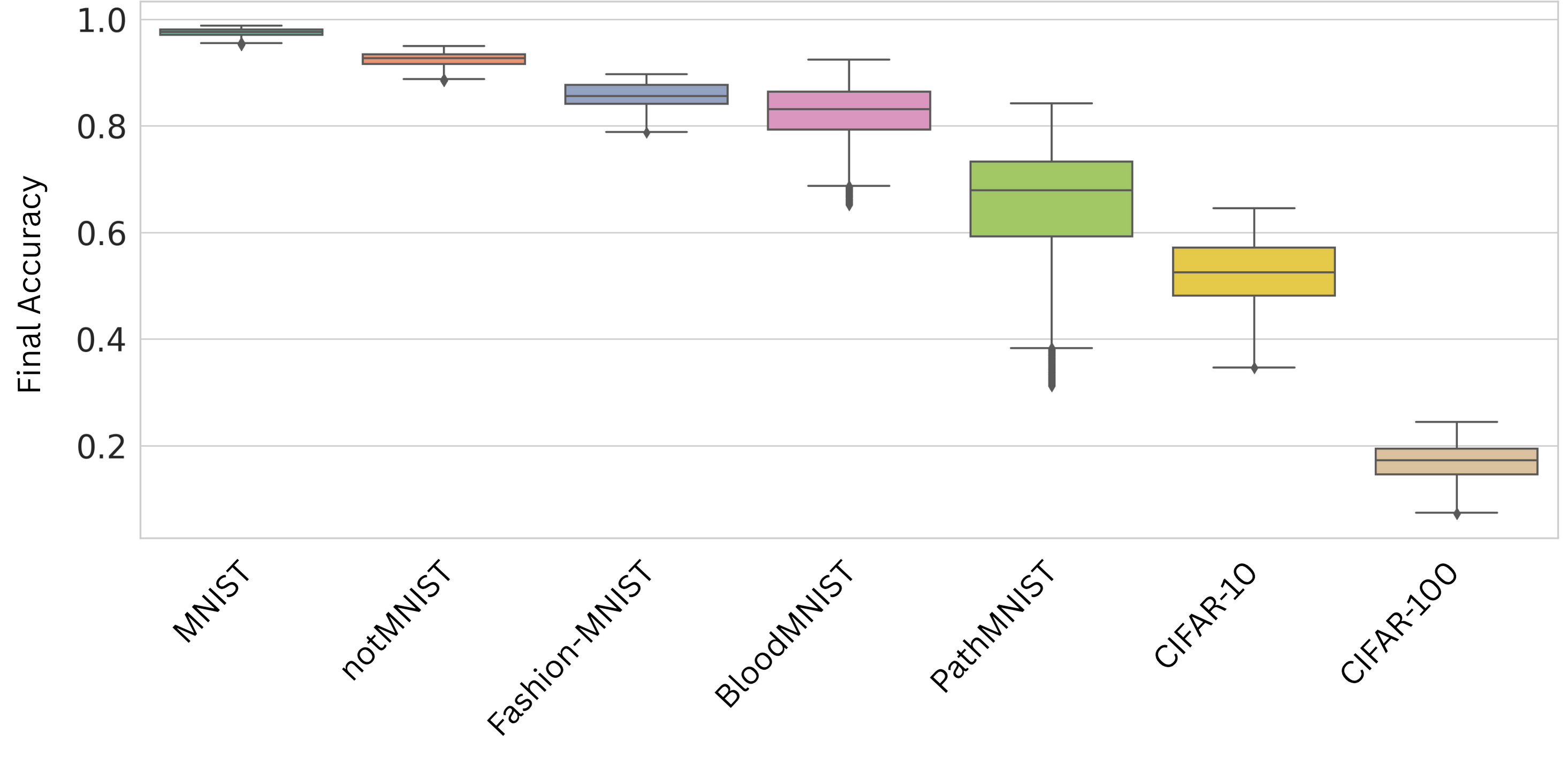}
  \caption{Distribution of model accuracy across datasets, highlighting the influence of dataset complexity.}
  \label{fig:accuracy_distribution}
\end{figure}

As shown in Figure~\ref{fig:accuracy_distribution}, simpler datasets (e.g., MNIST) yield consistently high accuracies with low variance, whereas complex datasets (e.g., CIFAR-100) exhibit lower and more variable accuracies. This pattern suggests that dataset difficulty is a dominant driver of performance, motivating Stage~1's linear baseline estimation and leaveing architecture-conditioned deviations to Stage~2.

\subsubsection{Latent Axes of Complexity: A PCA-Based Analysis}
\label{sec:pca_analysis}
We apply PCA to standardized DCMs to address multicollinearity and extract compact complexity bases. Standardization and PCA are fit only on the training folds in each cross-validation split to prevent leakage; the learned transforms are then applied to the held-out dataset.

\paragraph{Selection of Principal Components}

\begin{figure}[htbp]
    \centering
    \includegraphics[width=0.9\linewidth]{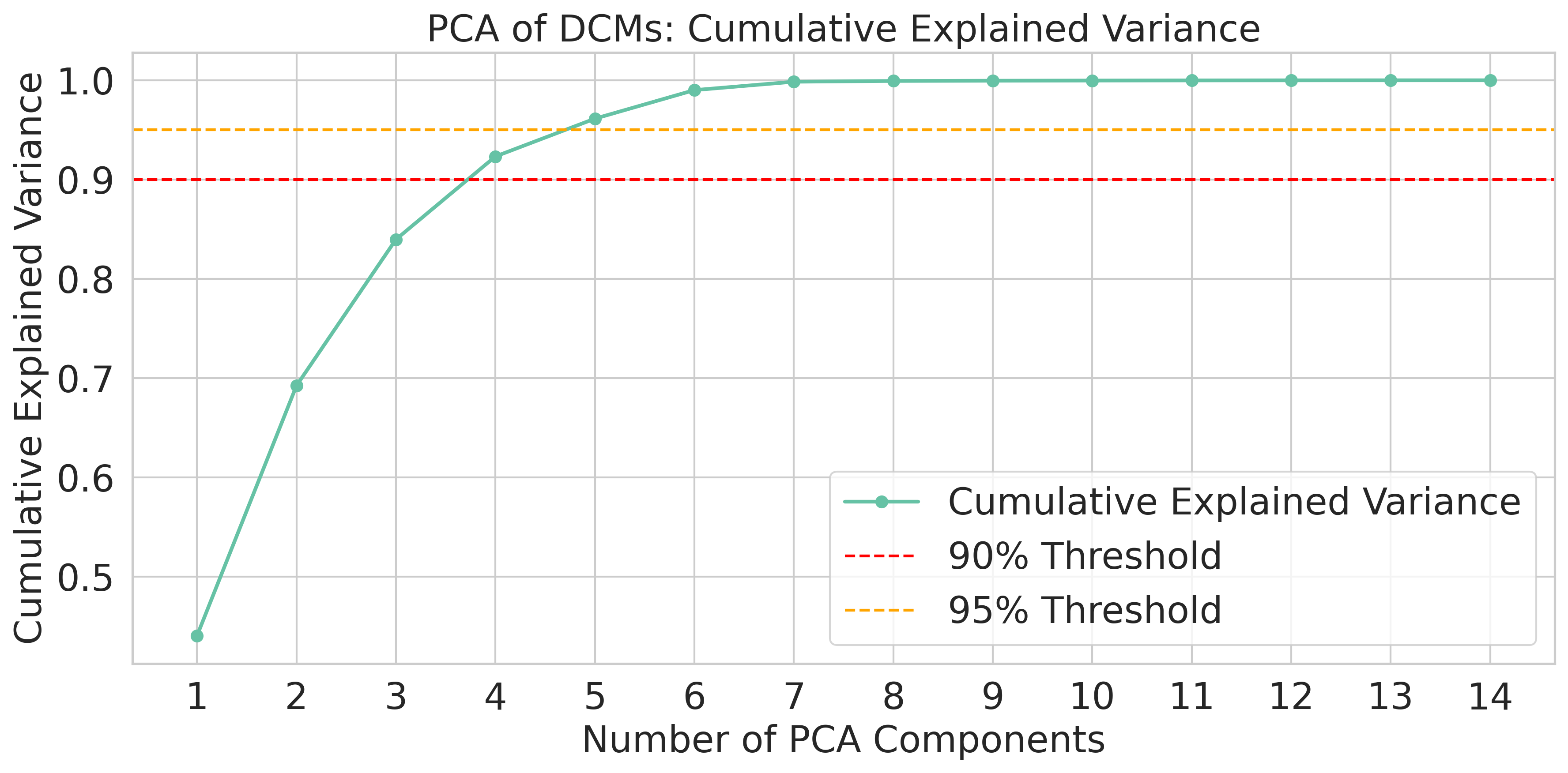}
    \caption{Explained variance ratio as a function of the number of principal components.}
    \label{fig:pca_variance}
\end{figure}

\begin{figure}[htbp]
    \centering
    \includegraphics[width=0.95\linewidth]{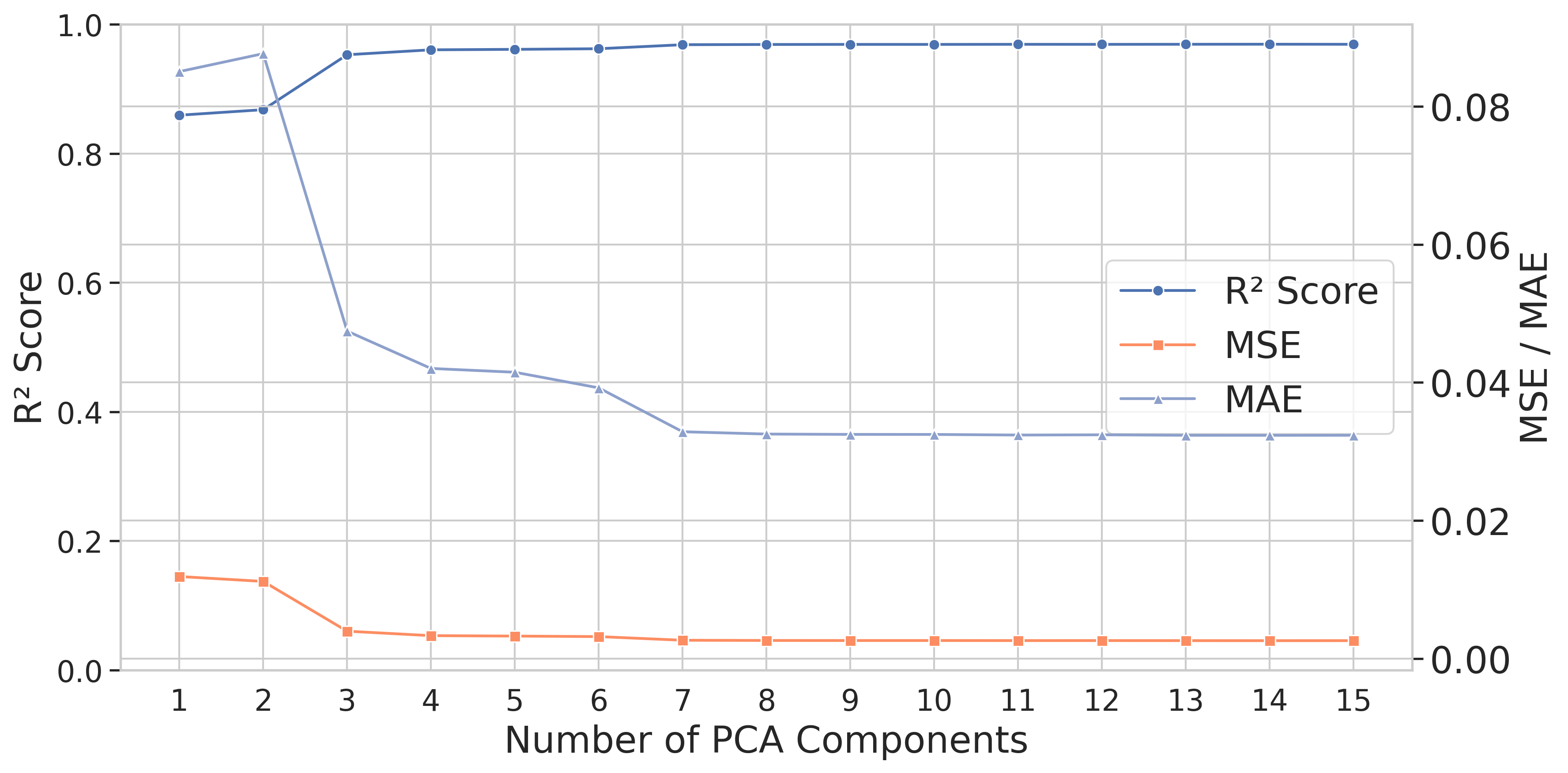} 
    \caption{Regression performance vs. number of PCA components. Metrics include $R^2$, MSE, and MAE. The best performance is achieved at $N = 7$.}
    \label{fig:pca_model_eval}
\end{figure}

Figure~\ref{fig:pca_variance} shows that the top four components explain \(\approx90\%\) of the variance, and seven components exceed \(95\%\). Regression performance for predicting dataset-level accuracy is maximized at \(N=7\) (Figure~\ref{fig:pca_model_eval}); we thus retain seven components downstream.

In addition, we evaluated linear regression performance on predicting dataset-level accuracy using different numbers of components. As illustrated in Figure~\ref{fig:pca_model_eval}, metrics such as MSE and MAE reached their lowest values at $N = 7$, confirming it as the optimal balance point between model complexity and predictive accuracy. Therefore, we used the top seven principal components for all downstream modeling.

\paragraph{Interpretation of PC1: Local Non-Linear Entanglement}
PC1 primarily captures localized, non-linear complexity. It is characterized by high positive loadings from neighborhood-based metrics such as NN Non-linearity ,NN Distance Ratio, and kNN Error Rate. Covariance Mean also contributes moderately. These reflect how irregularly classes are distributed in local regions of the feature space.

High PC1 values indicate a violation of the local smoothness assumption, where neighboring data points frequently belong to different classes. In such scenarios, deep models tend to demand higher representational capacity, stronger regularization, and more sophisticated optimization procedures to generalize effectively.

\begin{figure}[htbp]
    \centering
    \includegraphics[width=0.9\linewidth]{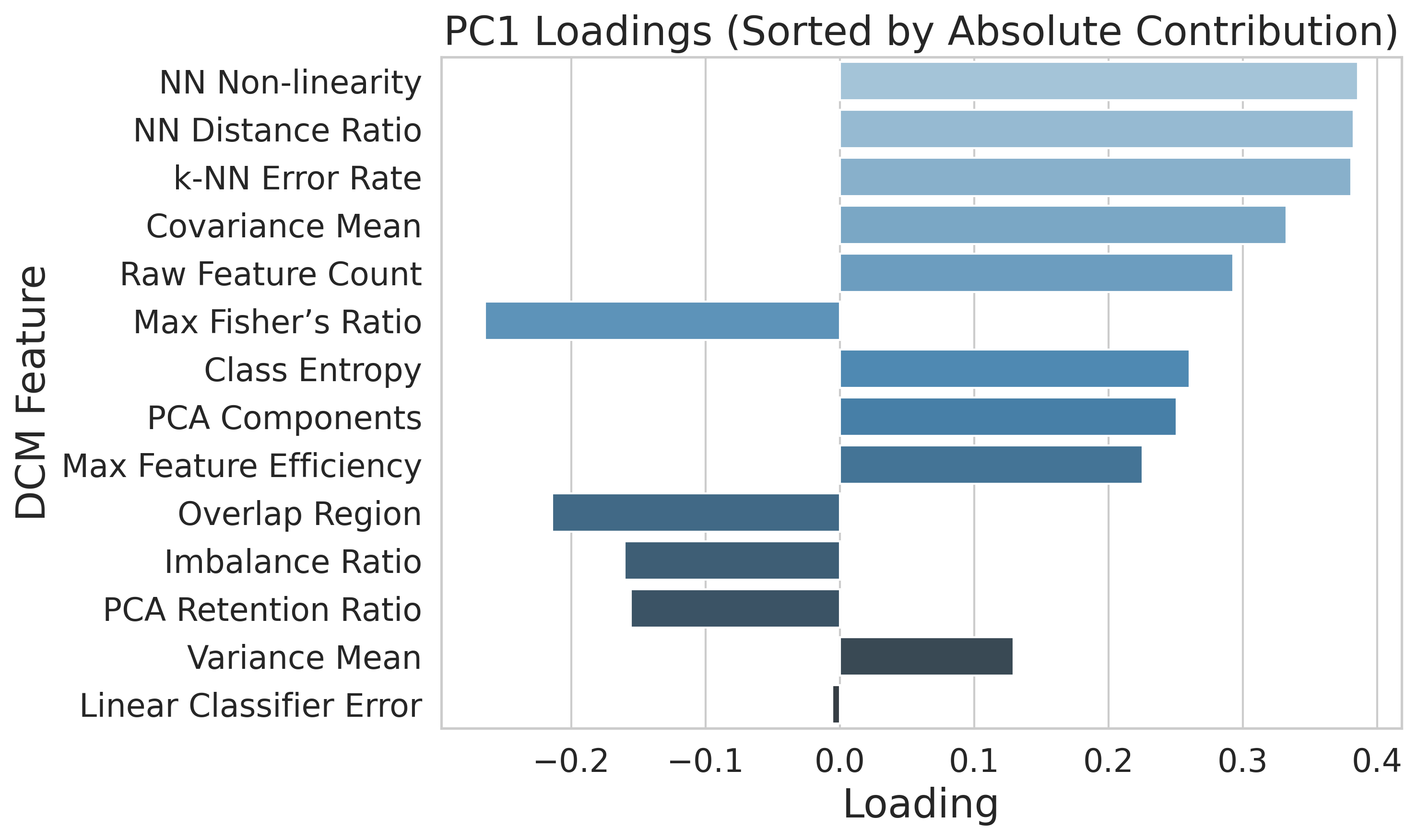}
    \caption{Loadings of DCMs on PC1: Local non-linear entanglement.}
    \label{fig:pc1_loadings}
\end{figure}

\paragraph{Interpretation of PC2: Global Structure and Feature Scale}
PC2 reflects global linear separability and feature scale. It has a strong negative loading from Linear Classifier Error Rate and positive loading from Variance Mean. High PC2 values suggest datasets with linearly separable class structures and high intra-class variation—conditions under which deep models can learn more effectively.

Datasets with high PC2 also tend to be more sensitive to training conditions (e.g., learning rate, weight initialization), and benefit from normalization techniques such as Batch Normalization.

\begin{figure}[htbp]
    \centering
    \includegraphics[width=0.9\linewidth]{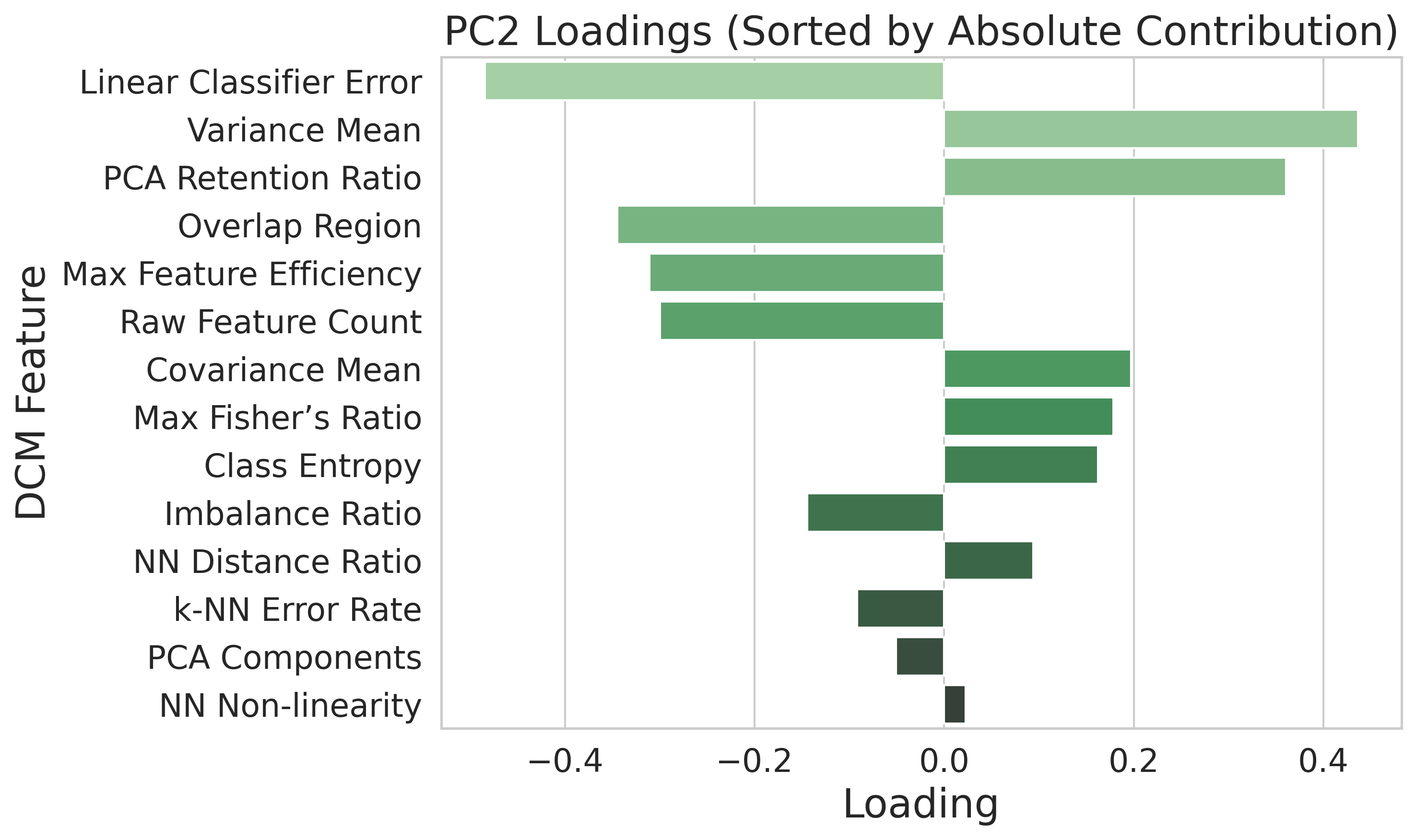}
    \caption{Loadings of DCMs on PC2: Global separability and feature scale.}
    \label{fig:pc2_loadings}
\end{figure}


\subsubsection{Synthesis: Data-Driven Rationale for the Two-Stage Framework}

The preceding analyses reveal two key insights: 1) Mean model performance is largely a linear function of the dataset itself, and 2) The complexity features that describe the dataset are organized along two distinct axes: local, non-linear entanglement (PC1) and global, linear separability (PC2).

The proposed two-stage framework is motivated by the distinct structure revealed in the data. Stage~1 captures global performance trends using a linear model, while Stage~2 accounts for non-linear variations driven by local entanglement (PC1), using a non-linear model such as XGBoost.


\subsection{Performance Validation of the Two-Stage Framework}

\subsubsection{Baseline Prediction Performance (Stage~1)}

To evaluate the ability of dataset-level complexity features to predict baseline performance, we applied both Linear Regression and Ridge Regression on the principal components of DCMs. These models were evaluated using Leave-One-Dataset-Out Cross-Validation (LODO-CV). The goal of Stage~1 is to estimate the expected model accuracy for a dataset, independent of architecture or hyperparameter configuration.

\paragraph{Regression Accuracy}
\label{sec:stage1_lodo_discussion}
Table~\ref{tab:stage1_lodo} summarizes the MAE, MSE, and $R^2$ for both models. Despite negative $R^2$ values (which are expected given the lack of configuration-level inputs), MAE remains low across all datasets. For instance, MAE ranges from $0.006$ (MNIST) to $0.088$ (PathMNIST), demonstrating that the DCM features are effective in capturing dataset difficulty.

\begin{table}[ht]
\centering
\caption{Stage~1 Baseline Prediction Performance under LODO-CV}
\label{tab:stage1_lodo}
\resizebox{\linewidth}{!}{%
\begin{tabular}{lllll}
\toprule
\textbf{Dataset} & \textbf{Model} & $R^2$ & \textbf{MSE} & \textbf{MAE} \\
\midrule
BloodMNIST & Linear & $-0.0017$ & $0.00283$ & $0.0428$ \\
           & Ridge  & $-0.0018$ & $0.00283$ & $0.0428$ \\
CIFAR-10   & Linear & $-0.0111$ & $0.00295$ & $0.0460$ \\
           & Ridge  & $-0.0109$ & $0.00295$ & $0.0460$ \\
CIFAR-100  & Linear & $-2.5073$ & $0.00397$ & $0.0536$ \\
           & Ridge  & $-2.8702$ & $0.00438$ & $0.0571$ \\
FashionMNIST & Linear & $-0.0784$ & $0.00048$ & $0.0183$ \\
           & Ridge  & $-0.0786$ & $0.00048$ & $0.0183$ \\
MNIST      & Linear & $-0.2897$ & $0.00007$ & $0.0063$ \\
           & Ridge  & $-0.2890$ & $0.00007$ & $0.0063$ \\
notMNIST   & Linear & $-2.3380$ & $0.00059$ & $0.0199$ \\
           & Ridge  & $-2.3395$ & $0.00059$ & $0.0199$ \\
PathMNIST  & Linear & $-0.1438$ & $0.01530$ & $0.0886$ \\
           & Ridge  & $-0.1394$ & $0.01524$ & $0.0885$ \\
\bottomrule
\end{tabular}
}
\end{table}

\paragraph{Visualization and Interpretation}

\begin{figure}[ht]
\centering
\includegraphics[width=0.95\linewidth]{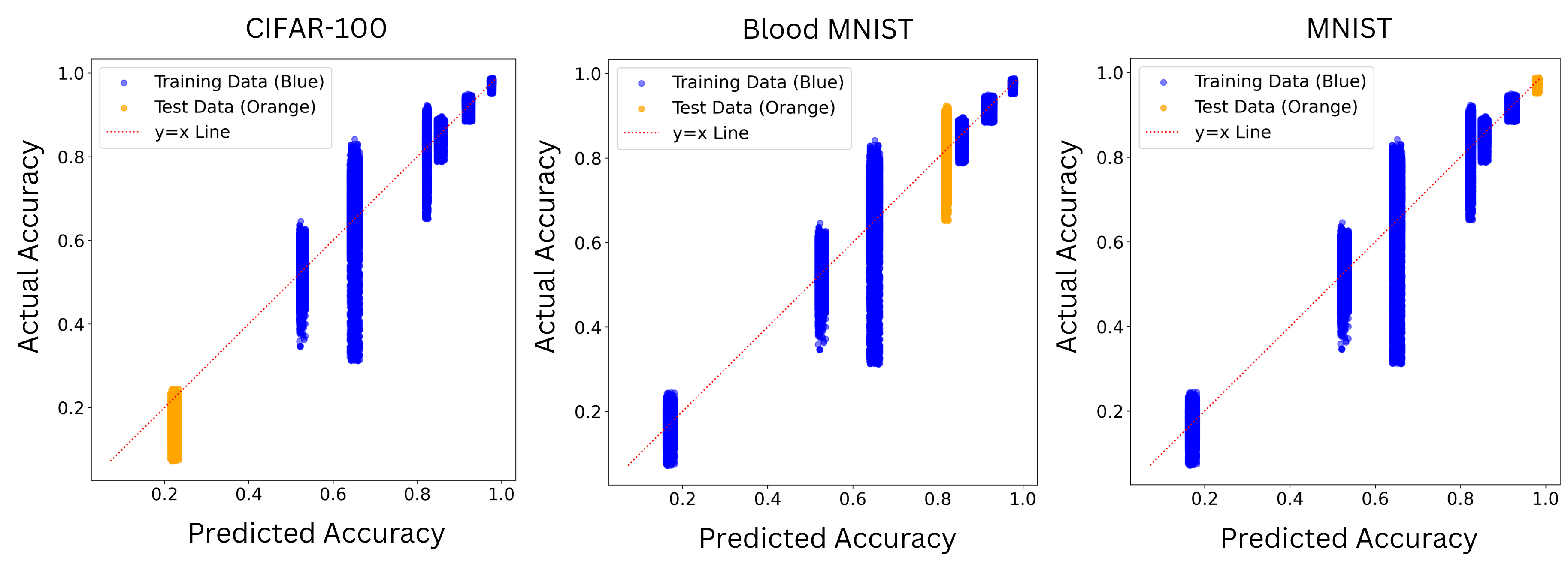}
\caption{Baseline prediction error (Stage~1) under LODO-CV using Linear Regression. Orange points indicate the held-out dataset.}
\label{fig:stage1_lodo}
\end{figure}

Figure~\ref{fig:stage1_lodo} illustrates the prediction error for each dataset under LODO-CV. Each orange point represents the held-out dataset prediction error using Linear Regression. While $R^2$ remains low due to the absence of configuration-level input, the model captures the general trend of performance across datasets of varying complexity.

\paragraph{Insights}

Notably, the prediction error remains consistent across datasets with vastly different difficulty levels—from CIFAR-100 (low accuracy) to MNIST (high accuracy)—indicating the robustness of DCM features. The results confirm that PCA-compressed DCMs encode monotonic indicators of dataset difficulty that are effective across visual domains.

From a theoretical perspective, these findings align with bias–variance tradeoff theory. Stage~1 captures high-bias but low-variance estimations governed by data complexity. The remaining variance, arising from model-specific configurations, is addressed in Stage~2.

Moreover, the framework demonstrates strong cross-domain generalization. Baseline predictions maintain stable accuracy across handwriting, object, and medical datasets, suggesting that the DCM space is domain-invariant and structurally expressive.


\subsubsection{Full Model Prediction Performance}

This section presents the evaluation results of the full two-stage prediction framework, which combines dataset complexity features (Stage~1) with model configuration descriptors (Stage~2) to predict model test accuracy. We consider three evaluation settings: (1) in-distribution prediction, (2) Leave-One-Dataset-Out (LODO) cross-validation, and (3) cross-domain generalization (LODM). Stage~2 is tested using three nonlinear regressors: Random Forest, Gradient Boosting, and XGBoost.

\paragraph{In-Distribution Prediction}

\begin{figure}[ht]
\centering
\includegraphics[width=0.95\linewidth]{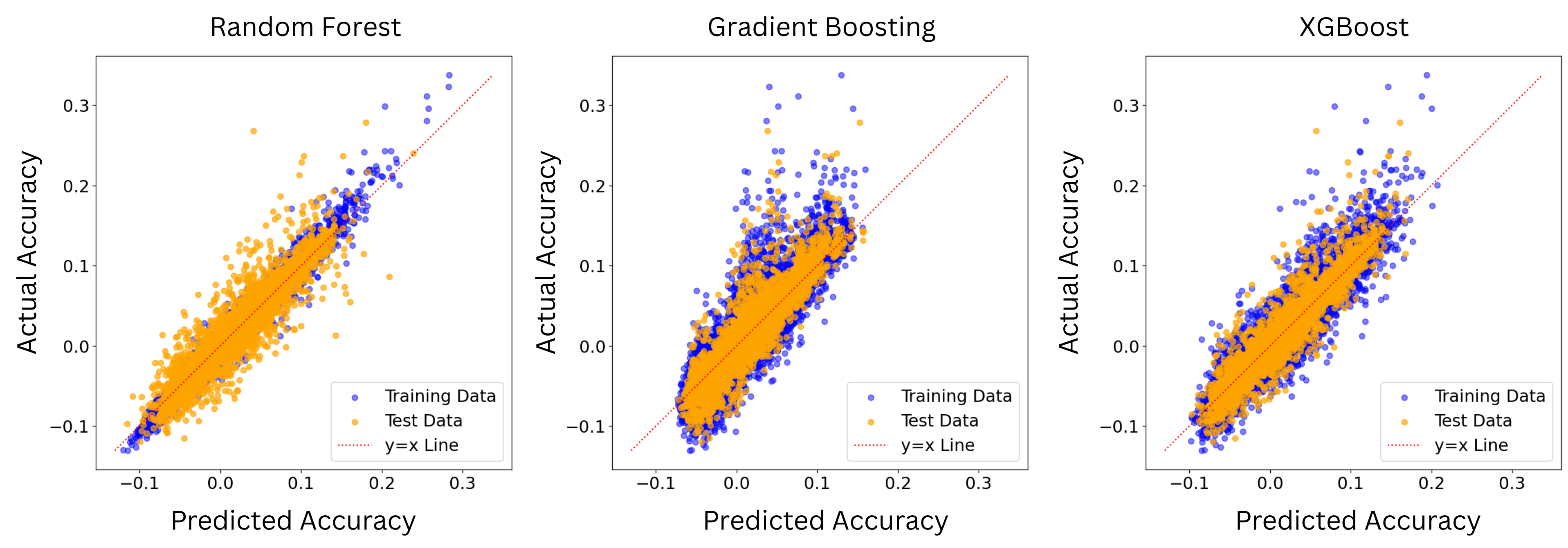}
\caption{Prediction error distribution on seen datasets (in-distribution). XGBoost yields the lowest average error.}
\label{fig:indist_error}
\end{figure}

We first evaluate the framework’s performance on configurations drawn from datasets included in training. This provides an upper bound reference for achievable accuracy and confirms the expressive capacity of Stage~2 models. Table~\ref{tab:stage2_indist} reports the $R^2$, MSE, and MAE for each regressor, while Figure~\ref{fig:indist_error} visualizes prediction errors.

\begin{table}[ht]
\centering
\caption{In-Distribution Prediction Performance (Stage~2)}
\label{tab:stage2_indist}
\begin{tabular}{lccc}
\toprule
\textbf{Model} & $R^2$ & \textbf{MSE} & \textbf{MAE} \\
\midrule
Random Forest     & 0.8144 & 0.000591 & 0.0124 \\
Gradient Boosting & 0.7604 & 0.000763 & 0.0165 \\
XGBoost           & \textbf{0.8206} & \textbf{0.000571} & \textbf{0.0120} \\
\bottomrule
\end{tabular}
\end{table}

All three models achieve high predictive accuracy, with XGBoost delivering the best performance. It explains over 82\% of the variance in accuracy and yields a mean absolute error below 1.3 percentage points. Random Forest performs comparably, while Gradient Boosting slightly underperforms, can be attributed to its sequential learning strategy, which may limit its expressiveness on highly nonlinear offset mappings.

These results confirm that ensemble-based nonlinear regressors effectively model the relationship between dataset-level complexity, architectural design, and configuration-induced performance shifts.

\paragraph{Leave-One-Dataset-Out Cross-Validation}

\begin{figure}[ht]
\centering
\includegraphics[width=0.95\linewidth]{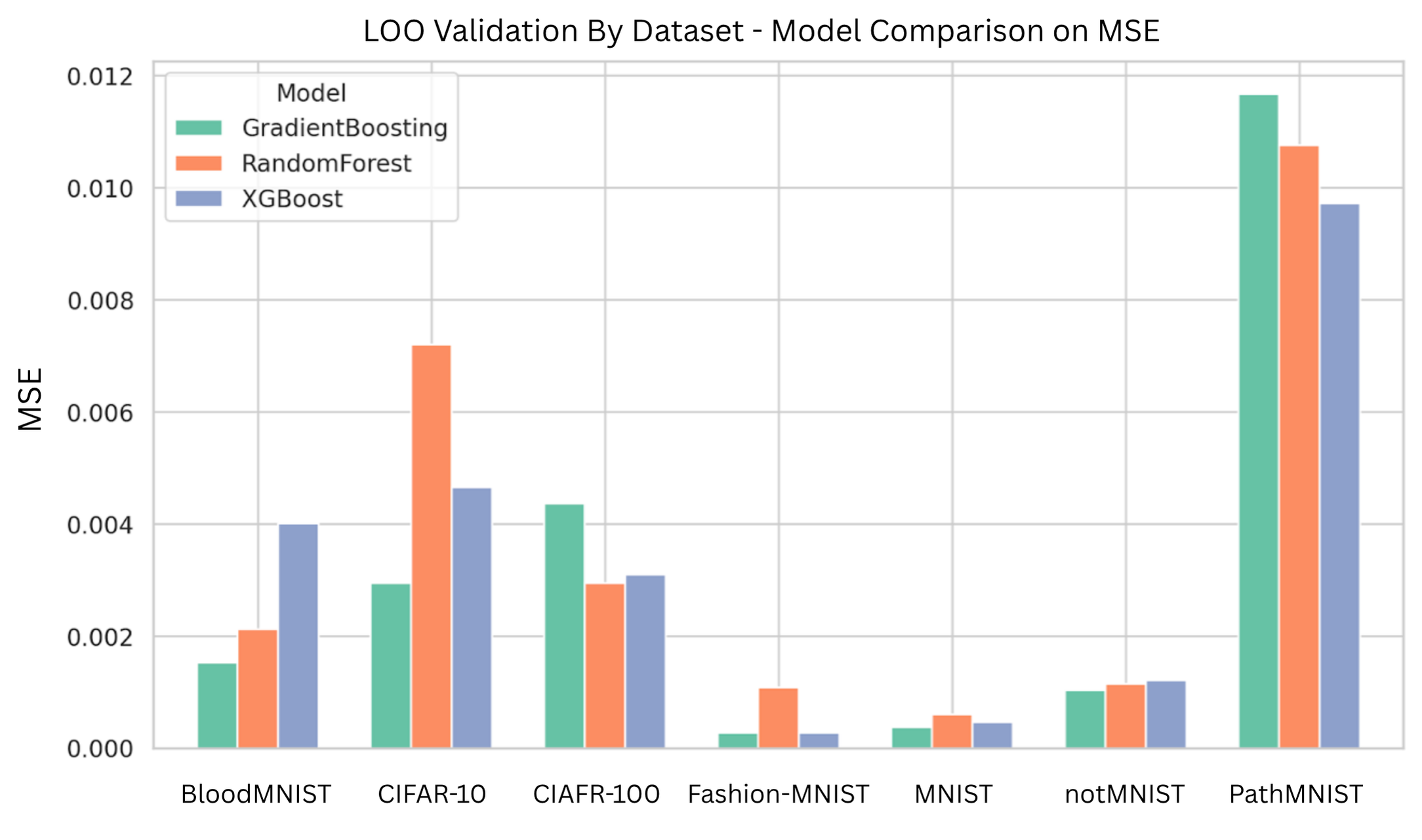}
\caption{Stage~2 LODO Cross-Validation: $R^2$, MSE, and MAE for each dataset-model pair.}
\label{fig:lodo_results}
\end{figure}

To assess the framework’s ability to generalize to entirely unseen datasets, we performed Leave-One-Dataset-Out (LODO) cross-validation. In each fold, one dataset was held out for testing while the two-stage model was trained on the remaining six. Table~\ref{tab:stage2_lodo} and Figure~\ref{fig:lodo_results} summarize the $R^2$, MSE, and MAE for Random Forest, Gradient Boosting, and XGBoost.

\begin{table}[ht]
\centering
\caption{Stage~2 LODO Cross-Validation Results}
\label{tab:stage2_lodo}
\resizebox{\linewidth}{!}{%
\begin{tabular}{llccc}
\toprule
\textbf{Test Dataset} & \textbf{Model} & $R^2$ & \textbf{MSE} & \textbf{MAE} \\
\midrule
BloodMNIST   & Random Forest     &  0.2444 & 0.002125 & 0.0370 \\
             & Gradient Boosting &  0.4554 & 0.001531 & 0.0297 \\
             & XGBoost           & -0.4270 & 0.004013 & 0.0556 \\
CIFAR-10     & Random Forest     & -1.4640 & 0.007205 & 0.0611 \\
             & Gradient Boosting & -0.0077 & 0.002947 & 0.0451 \\
             & XGBoost           & -0.5922 & 0.004656 & 0.0505 \\
CIFAR-100    & Random Forest     & -1.5741 & 0.002951 & 0.0419 \\
             & Gradient Boosting & -2.8130 & 0.004371 & 0.0569 \\
             & XGBoost           & -1.6865 & 0.003094 & 0.0436 \\
Fashion-MNIST& Random Forest     & -1.3302 & 0.001080 & 0.0256 \\
             & Gradient Boosting &  0.4038 & 0.000276 & 0.0135 \\
             & XGBoost           &  0.4102 & 0.000273 & 0.0133 \\
MNIST        & Random Forest     & -9.5344 & 0.000598 & 0.0173 \\
             & Gradient Boosting & -5.4800 & 0.000368 & 0.0161 \\
             & XGBoost           & -7.2482 & 0.000468 & 0.0169 \\
notMNIST     & Random Forest     & -4.2204 & 0.001152 & 0.0286 \\
             & Gradient Boosting & -3.7021 & 0.001037 & 0.0272 \\
             & XGBoost           & -4.4682 & 0.001206 & 0.0291 \\
PathMNIST    & Random Forest     &  0.1965 & 0.010763 & 0.0691 \\
             & Gradient Boosting &  0.1282 & 0.011679 & 0.0746 \\
             & XGBoost           &  0.2742 & 0.009722 & 0.0641 \\
\bottomrule
\end{tabular}%
}
\end{table}

Predictive performance varies across datasets. Fashion-MNIST and PathMNIST achieve positive $R^2$, indicating good offset modeling, whereas MNIST and notMNIST show negative $R^2$, reflecting difficulty in explaining within-dataset variance when fully held out. Nevertheless, all models maintain MAE below 0.06, suggesting practically useful predictions even under dataset shift.

Gradient Boosting often outperforms on BloodMNIST and CIFAR-10, likely due to its sequential residual-fitting mechanism that regularizes against dataset-specific noise.

\paragraph{Leave-One-Domain-Out Cross-Validation}

To evaluate the framework’s ability to generalize across visual domains, we conducted Leave-One-Domain-Out (LODM) cross-validation. In each fold, all configurations from one domain—handwritten digits, medical imaging, or object recognition—were excluded from training. This scenario tests whether the learned mapping from data complexity and model configuration to accuracy holds under substantial domain shift.

\begin{figure}[ht]
\centering
\includegraphics[width=0.95\linewidth]{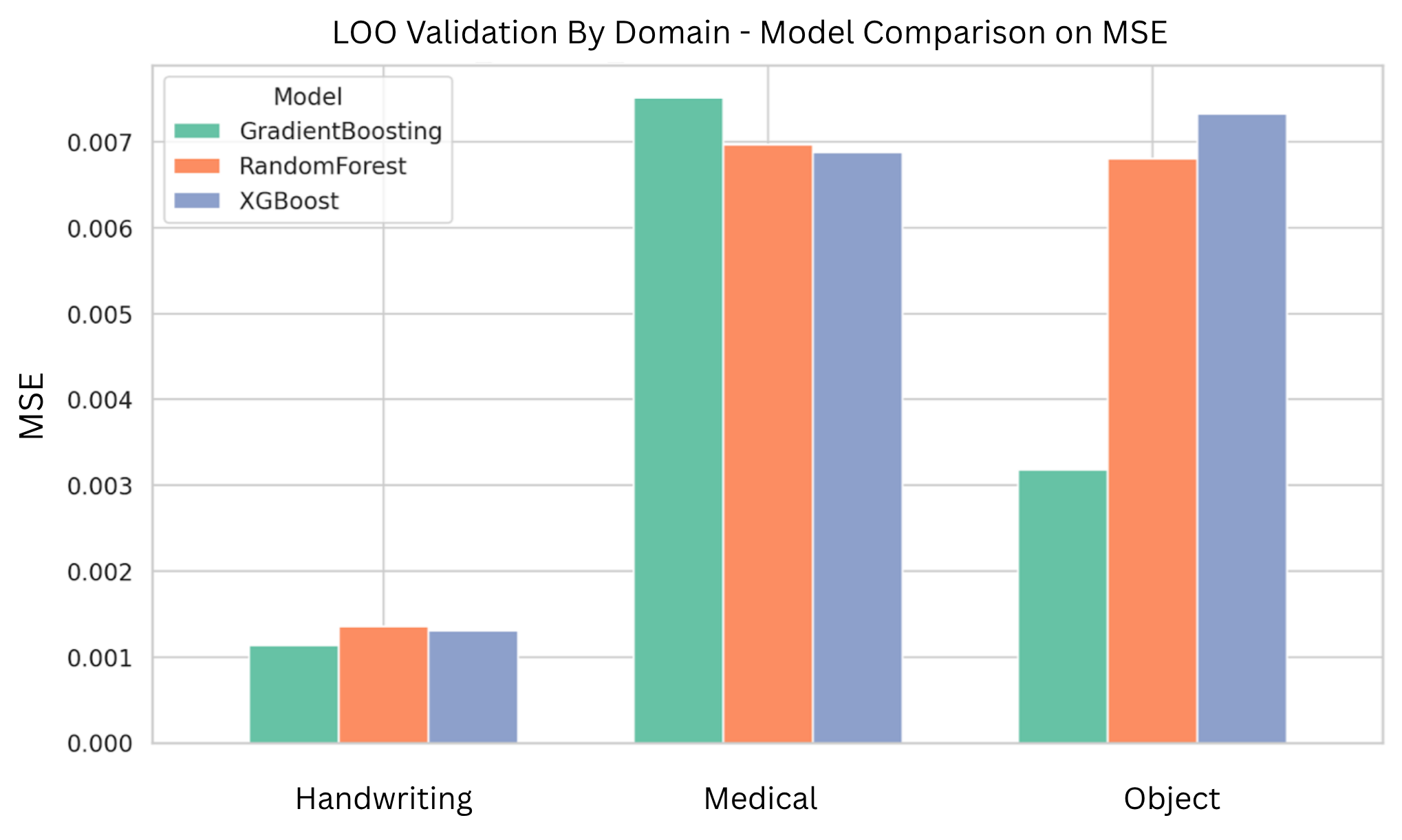}
\caption{Cross-domain prediction performance across handwritten, object, and medical domains.}
\label{fig:lodm_results}
\end{figure}

Table~\ref{tab:stage2_lodm} summarizes the results for the three Stage~2 regressors, and Figure~\ref{fig:lodm_results} visualizes their relative performance.

\begin{table}[ht]
\centering
\caption{Stage~2 Leave-One-Domain-Out Cross-Validation Results}
\label{tab:stage2_lodm}
\resizebox{\linewidth}{!}{%
\begin{tabular}{llccc}
\toprule
\textbf{Domain} & \textbf{Model} & $R^2$ & \textbf{MSE} & \textbf{MAE} \\
\midrule
Handwritten & Random Forest     & -5.4637 & 0.001357 & 0.0273 \\
            & Gradient Boosting & -4.4084 & 0.001136 & 0.0285 \\
            & XGBoost           & -5.2267 & 0.001307 & 0.0286 \\
Medical     & Random Forest     &  0.2346 & 0.006971 & 0.0554 \\
            & Gradient Boosting &  0.1749 & 0.007515 & 0.0555 \\
            & XGBoost           &  0.2450 & 0.006876 & 0.0510 \\
Object      & Random Forest     & -1.9326 & 0.006806 & 0.0586 \\
            & Gradient Boosting & -0.3715 & 0.003183 & 0.0419 \\
            & XGBoost           & -2.1568 & 0.007326 & 0.0571 \\
\bottomrule
\end{tabular}
}
\end{table}

\begin{figure}[htbp]
\centering
\includegraphics[width=0.95\linewidth]{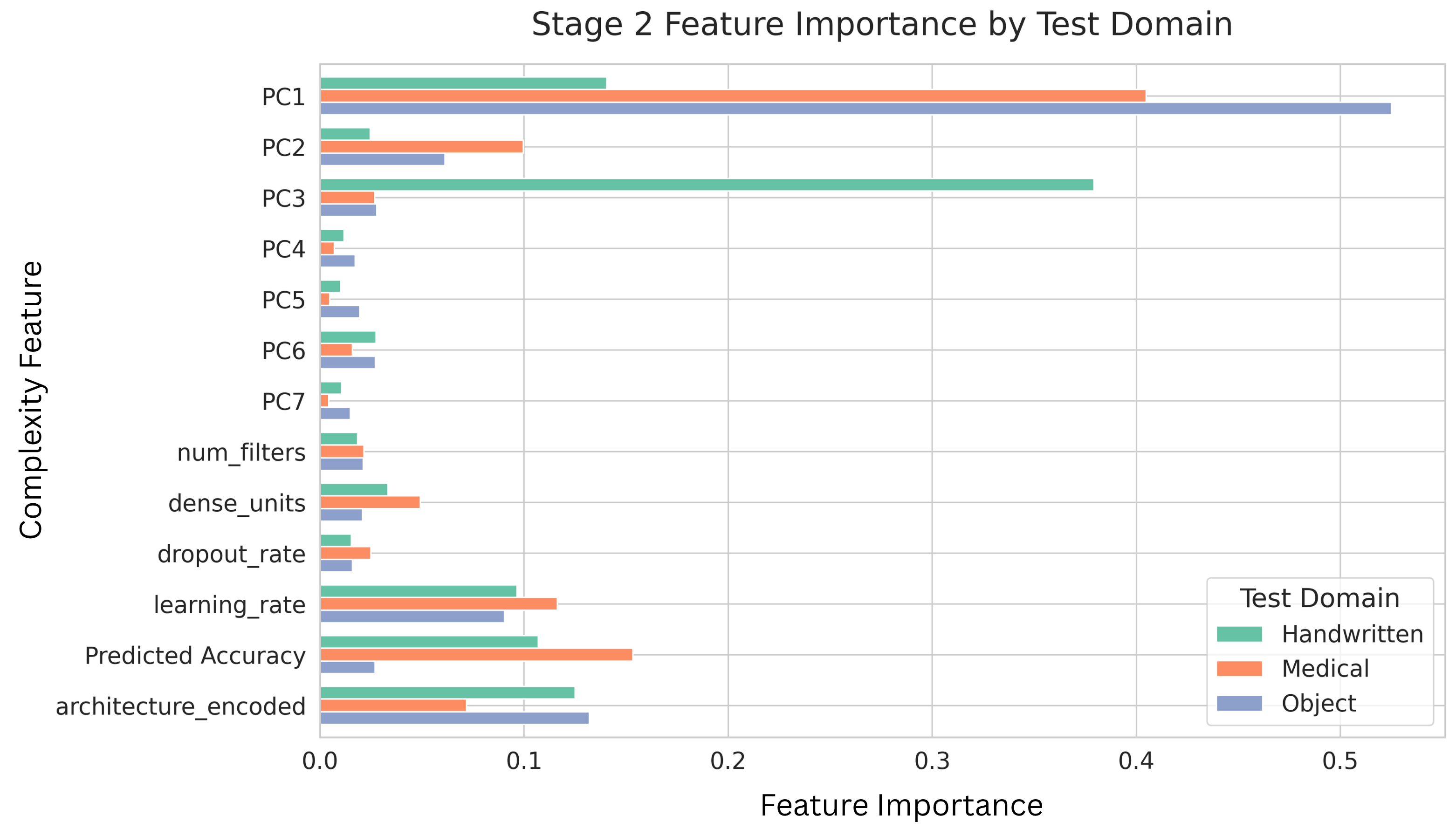}
\caption{Feature importance from Gradient Boosting model across domains. PC3 dominates prediction in the handwritten domain.}
\label{fig:feature_importance}
\end{figure}

Medical imaging tasks yielded the most consistent and positive $R^2$ across all models, indicating that dataset complexity metrics remained predictive even under domain shift. In contrast, handwritten digit tasks exhibited negative $R^2$, suggesting that while point estimates (MAE $\approx$ 0.028) remained reasonable, the models failed to capture within-domain variance when deprived of similar examples during training.

Object recognition results were mixed, with Gradient Boosting outperforming others in this domain, achieving the lowest MSE and MAE. This indicates its relative robustness to the higher statistical diversity typical of natural image datasets.

To further investigate domain-level effects, we analyzed feature importance scores from the Stage~2 Gradient Boosting model. As shown in Figure~\ref{fig:feature_importance}, the handwritten domain relied heavily on PC3, which corresponds to the PCA retention ratio. This aligns with the hypothesis that handwritten characters often reside on compact, low-dimensional manifolds, making dimensionality a dominant signal.


\subsection{Ablation Study: Justifying the Framework's Design}
\label{sec:ablation_study}

The preceding validation confirms the robustness and generalization capability of our proposed framework across datasets and domains. However, it remains unclear whether the explicit two-stage design—separating baseline estimation from offset prediction—is essential for this robust generalization. To rigorously evaluate the necessity of our architectural design choice, we next conduct an ablation study. This study specifically compares our proposed two-stage model against a simpler, single-stage model that directly predicts the final accuracy without the decoupling strategy.

To validate the architectural design of the proposed framework, we conducted an ablation study comparing the full two-stage pipeline against single-stage variants. Each variant retains the same feature inputs but alters the prediction structure, enabling direct attribution of performance gains to the decomposition into baseline and offset stages.

The single-stage model utilizes the same Gradient Boosting regressor but is trained directly on the complete feature set (all PCA-transformed DCMs and architecture descriptors) to predict the final accuracy in one step. Our hypothesis is that this simpler model will struggle to generalize, as it must learn the global, linear trends of dataset difficulty and the complex, non-linear interaction effects simultaneously.

The performance of both models, evaluated under the rigorous Leave-One-Dataset-Out (LODO) cross-validation protocol, is summarized in Table~\ref{tab:ablation_study}.

\begin{table}[ht]
    \centering
    \small
    \setlength{\tabcolsep}{3pt} %
    
    \caption{Ablation Study: Performance Comparison of Single-Stage vs. Two-Stage Frameworks under LODO-CV.}
    \label{tab:ablation_study}
    \begin{tabular}{l ccc ccc}
        \toprule
        & \multicolumn{3}{c}{\textbf{Single-Stage}} & \multicolumn{3}{c}{\textbf{Two-Stage (Ours)}} \\ 
        \cmidrule(lr){2-4} \cmidrule(lr){5-7}
        \textbf{Dataset} & $\mathbf{R^2}$ & \textbf{MSE} & \textbf{MAE} & $\mathbf{R^2}$ & \textbf{MSE} & \textbf{MAE} \\
        \midrule
        BloodMNIST      & -0.17  & .0033 & .0427 & \textbf{0.46}  & \textbf{.0015} & \textbf{.0297} \\
        CIFAR-10        & -4.59  & .0163 & .1172 & \textbf{-0.01} & \textbf{.0029} & \textbf{.0451} \\
        CIFAR-100       & -103.9 & .1186 & .3419 & \textbf{-2.81} & \textbf{.0044} & \textbf{.0569} \\
        Fashion-MNIST   & -30.6  & .0142 & .1179 & \textbf{0.40}  & \textbf{.0003} & \textbf{.0135} \\
        MNIST           & -56.2  & .0031 & .0536 & \textbf{-5.48} & \textbf{.0004} & \textbf{.0161} \\
        notMNIST        & -15.2  & .0029 & .0502 & \textbf{-3.70} & \textbf{.0010} & \textbf{.0272} \\
        PathMNIST       & -1.39  & .0319 & .1601 & \textbf{0.13}  & \textbf{.0117} & \textbf{.0746} \\
        \bottomrule
    \end{tabular}
\end{table}

The results unequivocally demonstrate the superiority of the two-stage design. The single-stage model consistently fails to generalize, yielding large, negative $R^2$ values across all datasets. This indicates its predictions are significantly worse than a naive model that simply predicts the mean accuracy. In contrast, our proposed two-stage framework achieves substantially lower Mean Squared Error (MSE) and Mean Absolute Error (MAE) in every single case. For instance, on Fashion-MNIST, the two-stage design reduces MAE by nearly 90\% and achieves a positive $R^2$, showcasing its robust predictive capability.

This performance gap supports our hypothesis. The single-stage model underperforms because it must simultaneously learn two distinct patterns—a stable global linear trend and a complex local non-linear interaction—using the same feature set. This conflation of objectives limits its generalization capacity. In contrast, the two-stage framework separates these tasks: the first stage uses a linear model to capture the global baseline, while the second stage employs a gradient boosting model to estimate the non-linear offset.

Therefore, this ablation study provides strong empirical evidence that the two-stage architecture is not an incremental improvement but a \emph{necessary design choice} for achieving robust and generalizable performance prediction.


\subsection{Deeper Insights: From Prediction to Guidance}
\label{sec:insights_guidance}

Beyond predictive accuracy, an additional contribution of our framework lies in its ability to provide interpretable insights that can inform practical MLOps decisions. In particular, we address a commonly encountered question in practice: \textit{How can one select an appropriate model architecture for a given dataset without extensive trial-and-error?} Our analysis suggests that \textit{Variance Mean}, one of the proposed complexity measures, may serve as a useful data-driven indicator for guiding model depth selection.

To examine this, we analyzed which dataset characteristics are most associated with architecture-dependent performance variations. Specifically, we computed the standard deviation of performance offsets across model configurations and applied PCA to the resulting variability profiles. As shown in Table~\ref{tab:pca_offset_variability}, Variance Mean exhibits the highest absolute loading, indicating its strong association with performance variability across architectures.

\begin{table}[htbp]
\centering
\caption{Principal Component Loadings on Offset Variability. Variance Mean shows the strongest correlation with performance variation.}
\label{tab:pca_offset_variability}
\begin{tabular}{llr}
\toprule
\textbf{Rank} & \textbf{Complexity Metric} & \textbf{Loading} \\
\midrule
1 & \textit{Variance Mean} & -0.80 \\
2 & Avg. number of PCA dimensions & +0.62 \\
3 & Maximum individual feature efficiency & +0.59 \\
4 & Error rate of linear classifier & +0.57 \\
5 & Error rate of NN classifier & +0.44 \\
\bottomrule
\end{tabular}
\end{table}

Having identified Variance Mean as the key driver, we conducted a two-way ANOVA to formally test its interaction with model depth. The results in Table~\ref{tab:anova_variance_depth} confirm a highly significant interaction effect ($p < 0.001$). This provides definitive statistical evidence that the optimal model depth is not a fixed choice, but is conditional upon the dataset's feature dispersion, as quantified by Variance Mean.

\begin{table}[htbp]
\centering
\caption{Two-Way ANOVA on Variance Mean and Model Depth.}
\label{tab:anova_variance_depth}
\begin{tabular}{lrr}
\toprule
\textbf{Source} & \textbf{F-value} & \textbf{p-value} \\
\midrule
Variance Mean (quintile) & 2902.17 & $< 0.001$ \\
Model Depth & 280.25 & $< 0.001$ \\
\textbf{Interaction (Variance $\times$ Depth)} & \textbf{89.30} & \textbf{$< 0.001$} \\
\bottomrule
\end{tabular}
\end{table}

\subsubsection{A Data-Driven Guideline for Architecture Selection}

This validated interaction yields a practical, data-driven principle for architecture design that can be applied before any model training:

\paragraph{For datasets with Low Variance Mean}
These datasets exhibit weakly dispersed features. To achieve high performance, deeper models are required. The intuition is that deeper architectures are better equipped to integrate and amplify subtle, low-energy signals scattered across multiple features. Simpler, shallower models are likely to underfit and fail to capture these nuanced patterns.

\paragraph{For datasets with High Variance Mean}
These datasets have widely distributed, high-energy features where class-separating information is more readily available. In this scenario, shallow or medium-depth models are sufficient and more efficient. Increasing model depth yields diminishing returns on accuracy while incurring significantly higher computational costs.

This analysis elevates our framework from a simple prediction tool to a prescriptive guide. By computing a single, lightweight metric (Variance Mean) pre-training, practitioners can make a more informed decision on architectural depth, thereby intelligently constraining the search space and accelerating the model development lifecycle.


\subsection{Practical Considerations and Diagnostic Capabilities}
\label{sec:practicality_and_diagnostics}

To support the framework’s integration into real-world MLOps workflows, this section discusses two practical aspects: its data efficiency and its diagnostic potential. We first evaluate the framework’s sensitivity to training sample size. Subsequently, we examine how its prediction errors can be reinterpreted as informative pre-training signals for assessing dataset quality.

\subsubsection{Practicality: Sample Size Requirements for Reliable Prediction}

A crucial question for any predictive tool is the amount of data required to generate reliable estimates. To evaluate the data efficiency of our framework, we analyzed its predictive performance when the input Data Complexity Measures (DCMs) were computed from increasingly larger, random subsets of each dataset.

\begin{figure}[htbp]
  \centering
\includegraphics[width=0.95\linewidth]{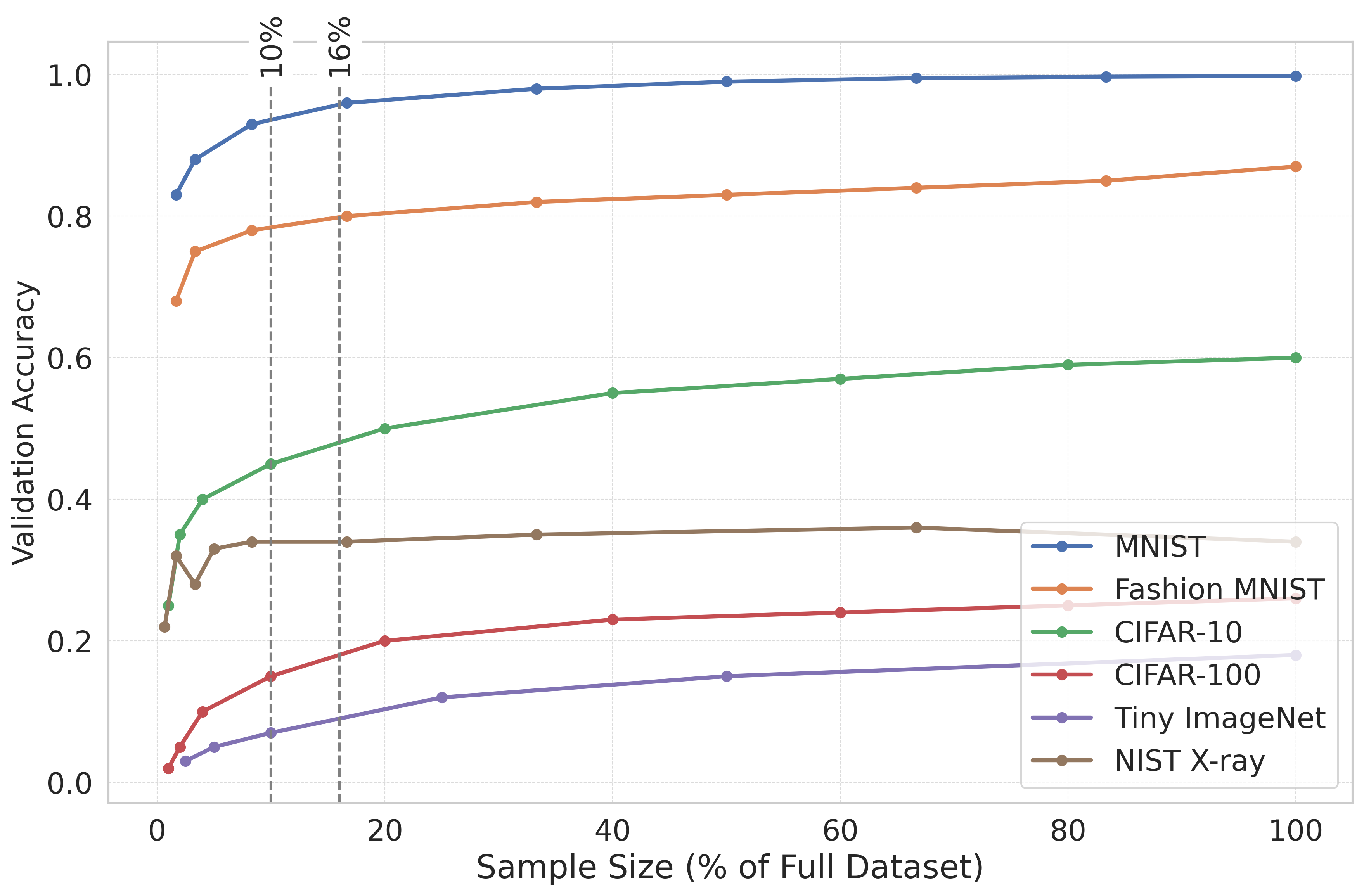}
  \caption{Framework prediction error (e.g., MSE) as a function of the data sampling ratio. Performance stabilizes and becomes reliable after approximately 16\% of the data is used to compute the DCMs.}
  \label{fig:sample_size_curve}
\end{figure}

Our findings, illustrated in Figure~\ref{fig:sample_size_curve}, indicate that the key data complexity characteristics stabilize early. The framework's prediction errors begin to plateau once the DCMs are computed on approximately 10\% of the total data. The predictive gains become marginal after sampling about 16\% of the data. 

This suggests that, for the purpose of performance prediction, only a fraction of the dataset is needed. We conclude that approximately 16\% of a dataset is generally sufficient to enable accurate predictions from our two-stage framework. This low sample size requirement significantly lowers the barrier to entry for applying our method, establishing it as a lightweight and efficient tool for rapid dataset assessment, especially in data-rich or time-sensitive environments.

\subsubsection{Diagnostic Utility: Interpreting Prediction Failures as a Data Quality Flag}
While the framework performs well on average, its behavior on outlier datasets is particularly insightful. In LODO cross-validation, we observed that prediction errors are not random; they are strongly correlated with a specific latent variable derived from the DCMs: the sixth principal component (PC6).

\paragraph{The Role of PC6 in Prediction Failures}
Analysis reveals a strong quadratic relationship between the Mean Squared Error (MSE) of our predictions and the PC6 score of the held-out dataset. Fitting a second-order polynomial yields the empirical relationship:
\begin{equation}
\mathrm{MSE} \approx -0.0008 + 0.0066 \cdot \mathrm{PC}_6 + 0.0159 \cdot \mathrm{PC}_6^2 \quad (R^2 = 0.84)
\end{equation}
This convex relationship, reinforced by the per-dataset error profiles in Table~\ref{tab:pc6_mse_table}, indicates a double-tailed risk: datasets with PC6 scores far from zero, either highly positive or negative, are prone to higher prediction errors.

\begin{table}[htbp]
\centering
\caption{Per-Dataset Prediction Error versus PC6 Score.}
\label{tab:pc6_mse_table}
\begin{tabular}{@{}lccc@{}}
\toprule
\textbf{Dataset} & $\mathbf{PC_6}$ & $\mathbf{|\mathrm{PC}_6|}$ & \textbf{MSE ($\times 10^{-3}$)} \\
\midrule
PathMNIST     & \phantom{-}0.611 & 0.611 & 10.7 \\
CIFAR-10      & -0.853 & 0.853 & 4.94 \\
CIFAR-100     & \phantom{-}0.379 & 0.379 & 3.47 \\
BloodMNIST    & -0.737 & 0.737 & 2.56 \\
notMNIST      & \phantom{-}0.345 & 0.345 & 1.13 \\
Fashion-MNIST & -0.195 & 0.195 & 0.54 \\
MNIST         & \phantom{-}0.188 & 0.188 & 0.48 \\
\bottomrule
\end{tabular}
\end{table}

\paragraph{Interpreting PC6 as a Dual-Risk Indicator}
PC6 primarily reflects a combination of two data characteristics: class distribution entropy and feature variance. That is,
\[
\mathrm{PC}_6 \text{ is largely composed of } \mathrm{ClassEntropy} \text{ and } \mathrm{VarianceMean}.
\]
Datasets with extreme PC6 values tend to exhibit either high class imbalance, high feature dispersion, or both—each posing distinct challenges to model training. Table~\ref{tab:pc6_risks} summarizes the associated risks.

\begin{table}[htbp]
\centering
\caption{Conceptual Risks Associated with Extreme PC6 Values.}
\label{tab:pc6_risks}
\begin{tabular}{@{}llll@{}}
\toprule
\textbf{PC6 Regime} & \textbf{Dominant Risk Profile} & \textbf{Example} \\
\midrule
$\mathrm{PC}_6 \gg 0$ & \textbf{Variance-dominated}: High class entropy & PathMNIST \\
(High & and high feature variance suggest noisy, & \\
positive) & heterogeneous inputs. & \\
\midrule
$\mathrm{PC}_6 \ll 0$ & \textbf{Bias-dominated}: Low class entropy & CIFAR-10 \\
(High & suggests significant class imbalance, & \\
negative) & leading to systematic bias. & \\
\midrule
$\mathrm{PC}_6 \approx 0$ & \textbf{Balanced Learning}: Moderate complexity & MNIST \\
(Near zero)& without extreme bias or variance. & \\
\bottomrule
\end{tabular}
\end{table}

\paragraph{Practical Implications: PC6 as a Pre-Training Heuristic}
This insight reinterprets a limitation of the framework as a potentially useful diagnostic heuristic. The PC6 score can be used as a pre-training indicator to inform data preprocessing strategies:
\begin{itemize}
    \item \em{If $\mathbf{PC_6 \gg 0}$ (Variance-dominated profile):} This suggests a high-variance dataset, possibly with label noise. Recommended actions include stratified sampling, noise-robust loss functions, or targeted data augmentation.
    \item \em{If $\mathbf{PC_6 \ll 0}$ (Bias-dominated profile):} This indicates potential class imbalance. Practitioners may consider class re-weighting, oversampling methods such as SMOTE, or other bias mitigation techniques.
\end{itemize}
In this way, the framework serves not only as a performance predictor but also as a meta-diagnostic tool, providing actionable signals for identifying and addressing dataset-specific risks prior to model training.

\subsection{Generalization Beyond Image Data}

Although our empirical evaluations primarily focus on image classification tasks, the proposed two-stage predictive framework is not inherently limited to the visual domain. Its broader applicability fundamentally hinges upon the use of Data Complexity Measures (DCMs), which quantify intrinsic dataset properties independent of specific input modalities~\cite{fulcher2013highly, gomaa2013survey}. Consequently, by adapting the feature extraction and complexity estimation process to alternative data modalities, our approach is theoretically capable of generalizing across structured, unstructured, and sequential data domains.

Specifically, extending this framework beyond image data involves defining or selecting appropriate complexity metrics tailored to the characteristics of the target modality. For tabular datasets, data complexity can be quantified using statistical descriptors analogous to image-based DCMs, including feature covariance, attribute sparsity, inter-class overlap, and intrinsic dimensionality~\cite{lorena2019complex}. These measures can be computed directly from raw numeric or categorical features without requiring model training, thus preserving computational efficiency.

In textual domains, complexity may be captured through linguistic and semantic metrics such as vocabulary entropy, lexical dispersion, and embedding-based semantic density~\cite{gomaa2013survey}. Such measures—rooted in corpus-based and knowledge-based similarity theory—allow estimation of dataset difficulty from pre-trained semantic spaces, enabling early-stage model selection and architecture design.

For sequential or time-series data, complexity can be estimated via dynamic characteristics including autocorrelation structure, entropy rate, and non-linearity indicators derived from state-space embeddings~\cite{fulcher2013highly}. These metrics reflect the underlying temporal variability and predictability of sequences, contributing to model selection and expected training dynamics.

The second stage of our framework, which models architecture-specific performance offsets, can similarly be generalized by adapting architectural descriptors to different deep model configurations. These descriptors may include network depth, width, kernel size, and other structural hyperparameters relevant to the target architecture class (e.g., convolutional layers in CNNs or transformer blocks). The core design—comprising baseline estimation via dataset-only properties and offset correction using model descriptors—thus remains structurally intact and conceptually valid across diverse data types and deep learning configurations.

Future work should empirically validate this generalization by benchmarking predictive accuracy and interpretability across textual, tabular, and sequential datasets, while applying appropriate deep architectures for each domain. Such investigations will further substantiate the methodological universality of the framework and expand its practical utility in cross-domain performance forecasting.


%% file: 500_conclusion.tex
\section{Conclusion}
\label{sec:conclusion_future_work}

\subsection{Conclusion}

This work establishes that deep model performance can be accurately predicted \textit{prior to training} by decoupling dataset difficulty from architecture-specific effects. We introduced a two-stage predictive framework that integrates data complexity measures (DCMs) with ensemble learning to provide generalizable, interpretable, and low-cost performance estimates. Our research delivers five key contributions:

\begin{itemize}
    \item {\em Feature Engineering with DCMs:} We show that data complexity measures, particularly the proposed \textit{Covariance Mean} and \textit{Variance Mean}, serve as compact and informative descriptors for predicting model accuracy.
    
    \item {\em Interpretability via PCA:} Principal Component Analysis (PCA) reduced correlated metrics into two interpretable axes of difficulty—{\em PC1} (local non-linear entanglement) and {\em PC2} (global structure)—which support the rationale for a two-stage framework.
    
    \item {\em Generalization under Distribution Shift:} The framework achieved low mean absolute error (MAE) under Leave-One-Dataset-Out (LODO) and Leave-One-Domain-Out (LODM) validation, indicating consistent performance on unseen datasets and domains.
    
    \item {\em Architecture Selection Heuristic:} Statistical analysis suggests that the \textit{Variance Mean} metric is associated with model depth, offering a practical pre-training heuristic for architecture selection.
    
    \item {\em Diagnostic Potential of PC6:} A prediction failure case was reinterpreted as a diagnostic signal. The latent component {\em PC6} may serve as an early indicator of datasets prone to bias or variance issues.
\end{itemize}

In summary, the proposed framework contributes to predictive MLOps by grounding model selection and data preprocessing in measurable dataset properties. This approach supports more systematic and interpretable deep learning workflows, potentially reducing reliance on resource-intensive trial-and-error procedures.

\subsection{Limitations}

Despite the promising results, we acknowledge several limitations that open avenues for future research.

\begin{itemize}
    \item {\em Scope of Architectures and Tasks:} The current evaluation focuses on convolutional architectures for image classification. The framework's generalization to other model families, such as Vision Transformers, and other tasks, including object detection or natural language processing, has not yet been assessed.
    
    \item {\em Handcrafted Nature of DCMs:} The feature set relies on manually designed data complexity measures, which may not fully capture the inductive biases or representational requirements of modern deep learning models.
    
    \item {\em Interpretability of Stage 2:} The XGBoost model used in Stage~2 offers limited interpretability, making it challenging to attribute predicted performance offsets to specific architectural or dataset characteristics.
\end{itemize}

